\documentclass[letterpaper, 10 pt, conference]{ieeeconf}

\IEEEoverridecommandlockouts
\usepackage{graphics}
\usepackage{epsfig}
\usepackage{mathptmx}
\usepackage{times}
\usepackage{amsmath}
\usepackage{amssymb}
\usepackage{mathtools}

\usepackage{graphicx}
\usepackage{subcaption}
\usepackage{placeins}
\usepackage{color} 
\usepackage{url}
\usepackage{stfloats}

\definecolor{somegray}{rgb}{0.5, 0.5, 0.5}
\newcommand{\darkgrayed}[1]{\textcolor{somegray}{#1}}
\makeatletter
\newcommand*\titleheader[1]{\gdef\@titleheader{#1}}
\AtBeginDocument{%
  \let\st@red@title\@title
  \def\@title{%
    \vskip-3em
    \bgroup\normalfont\large\centering\@titleheader\par\egroup
    \vskip1.5em\st@red@title}
}
\makeatother

\titleheader{\darkgrayed{This paper has been accepted for publication at the IEEE/RSJ International\\
Conference on Intelligent Robots and Systems (IROS), Abu Dhabi, 2024.
\copyright IEEE}}

\title{\LARGE \bf
Structure-Invariant Range-Visual-Inertial Odometry
}

\author{Ivan Alberico$^{1,2}$, Jeff Delaune$^{2}$, Giovanni Cioffi$^{1}$ and Davide Scaramuzza$^{1}$ \\ %
$^{1}$ Robotics and Perception Group, University of Zurich, Switzerland \\
$^{2}$ Jet Propulsion Laboratory, California Institute of Technology, USA
\thanks{}%
}

\begin{document}

\makeatletter
\g@addto@macro\@maketitle{
  \captionsetup{type=figure}\setcounter{figure}{0}
  \def\mycolspace{1.2mm}
   \centering
     \includegraphics[clip, width=1.99\columnwidth]{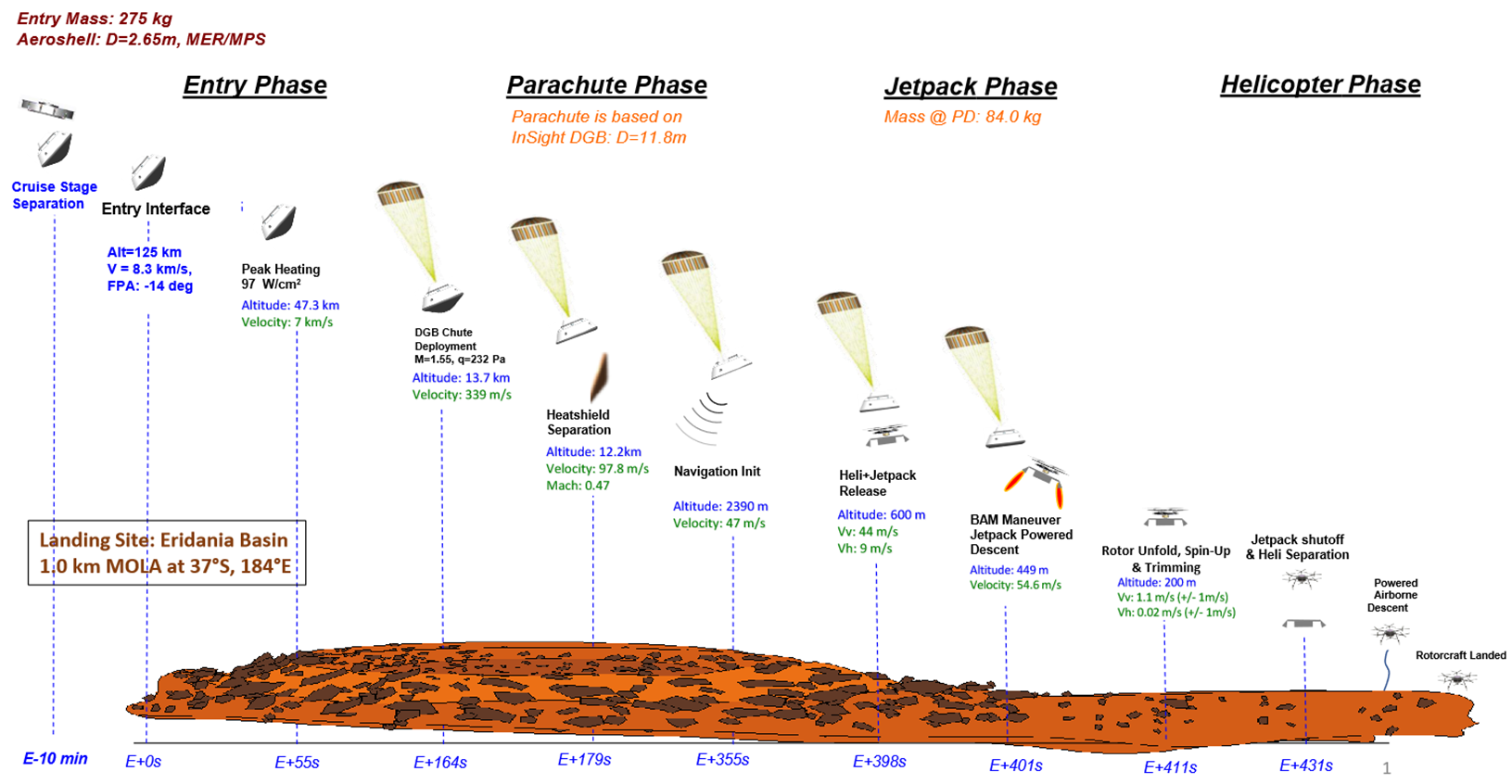}
     \vspace{1.0ex}
     \captionof{figure}{Concept of operations for Mid-Air Helicopter Delivery (MAHD). Values are specific to a national landing site in Mars' Eridania Basin. Velocity and altitude measurements are expressed with respect to the local terrain. The figure is extracted from our previous work \cite{9843825}.}
     \label{img:mahd_edl}
}
\makeatother
\maketitle
\thispagestyle{empty}
\pagestyle{empty}

\begin{abstract}
The Mars Science Helicopter (MSH) mission aims to deploy the next generation of unmanned helicopters on Mars, targeting landing sites in highly irregular terrain such as Valles Marineris, the largest canyons in the Solar system with elevation variances of up to 8000 meters.
Unlike its predecessor, the Mars 2020 mission, which relied on a state estimation system assuming planar terrain, MSH requires a novel approach due to the complex topography of the landing site.
This paper introduces a novel range-visual-inertial odometry system tailored for the unique challenges of the MSH mission. 
Our system extends the state-of-the-art xVIO framework by fusing consistent range information with visual and inertial measurements, preventing metric scale drift in the absence of visual-inertial excitation (mono camera and constant velocity descent), and enabling landing on any terrain structure, without requiring any planar terrain assumption.
Through extensive testing in image-based simulations using actual terrain structure and textures collected in Mars orbit, we demonstrate that our range-VIO approach estimates terrain-relative velocity meeting the stringent mission requirements, and outperforming existing methods.
\end{abstract}

\section*{Supplementary material}
\label{sec:supplementary_material}
A video of the proposed method is available at~\url{https://youtu.be/lp0Tu45YDIE}.

\section{INTRODUCTION}
\label{sec:introduction}

The Ingenuity Mars helicopter achieved the first autonomous flight on another planet during the Mars 2020 mission. Building on this success, the next Mars Science Helicopter (MSH) mission has been proposed to transport multiple kilograms of science payload over multiple kilometers~\cite{doi:10.2514/6.2020-4029}. 
The MSH navigation system has been designed to explore terrains previously deemed inaccessible to Ingenuity, such as cliffs, craters, and steep slopes~\cite{9438289}, offering new avenues for scientific exploration. 
A next-generation rotorcraft like the one proposed in MSH can enable scientists to explore large-scale areas on Mars~\cite{madh_geoloc}. 

MSH is a helicopter-only mission concept where the helicopter will be delivered mid-air after atmospheric entry and descent.
Mid-Air Helicopter Delivery (MAHD)~\cite{9843825} is a new Entry, Descent, and Landing (EDL) system using a jetpack to slow down the helicopter within its control envelope, as shown in Fig.~\ref{img:mahd_edl}.
The jetpack is controlled by helicopter avionics, including its vision-based navigation system. 

Since a helicopter already has all the functions to land, MAHD is expected to create volume margins, improve terrain access, and reduce cost by eliminating a traditional planetary lander. 
Vision-based navigation may start as high as 12000 m above the ground when the heatshield is jettisoned, and the helicopter camera can see the ground. 
During the jetpack phase, the vision-based navigation system provides position, velocity, and orientation feedback to bring the jetpack to a terrain-relative hover at 200 m above ground, where the helicopter then takes off. 
Velocity, orientation, and altitude estimates are critical for EDL control.
While orientation and altitude can be accurately estimated by integrating MSH gyroscopes from orbit and altimeter measurements, velocity estimates are challenging to achieve.
The range-VIO system, xVIO~\cite{DBLP:journals/ral/DelauneBB21}~\cite{DBLP:journals/corr/abs-2010-06677}, used in the Mars 2020 relied on terrain planar assumption.
In MHS, landing targets include areas with some of the most challenging 3D environments in the Solar system. 
For instance, in Valles Marineries, terrain elevation can vary as much as 8000 m in close proximity. 
This paper addresses the challenges of the MSH mission by introducing a method that eliminates the need for any type of ground planarity assumption, therefore adaptable to any terrain structure, while still being able to observe scale and mitigate error drift under constant-velocity motion and without relying on prior maps.

This paper claims three significant contributions:
\begin{enumerate}
    \item a novel range-VIO algorithm facilitating seamless integration of 1D-LRF measurements into the state-of-the-art framework, xVIO~\cite{DBLP:journals/ral/DelauneBB21}~\cite{DBLP:journals/corr/abs-2010-06677}, without relying on terrain planarity assumptions.
    \item a simulation environment specifically designed for MAHD testing. The environment accurately replicates the challenging conditions associated with MAHD, reproducing photorealistic 3D terrains with steep elevation slopes and rendered at high altitudes over constant velocity trajectories.
    \item an extensive assessment of the proposed algorithm in Monte Carlo analysis, showing superior performance compared to state-of-the-art range-VIO.
\end{enumerate}

\begin{figure}[t]
     \vspace{2mm}
     \centering
     \includegraphics[width=0.48\textwidth]{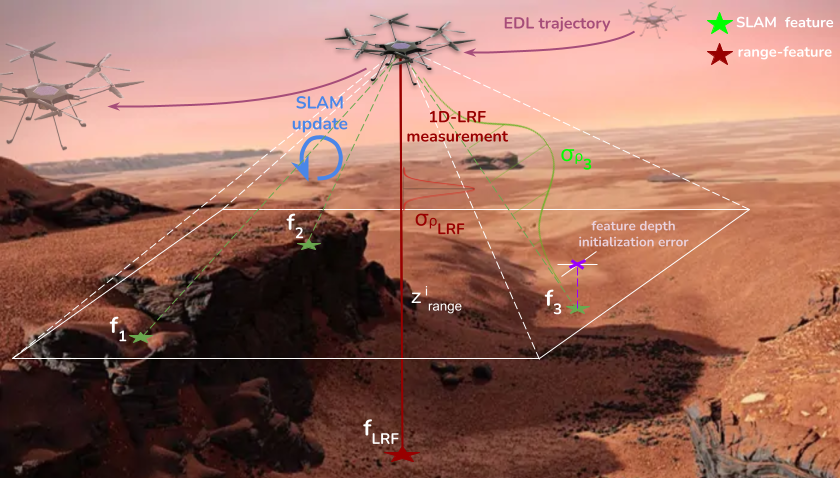}
     \caption{Schematic representation of the proposed method. The red star corresponds to a range-feature whose depth is initialized with the 1D-LRF measurement and standard deviation $\sigma_{\rho_{LRF}}$.}
     \label{img:representation_range_feature_initialization}
\end{figure}

\section{LITERATURE REVIEW}
\label{sec:literature_review}

Two bodies of literature are relevant for MAHD: aerial navigation on Mars and range-visual-inertial odometry in the robotics literature. While drones can fly with other sensors on Earth (GPS, radar, lidar), they are not an option for MSH.

\subsection{Vision-based Navigation for Mars}

State of flight for Mars aerial vision-based navigation is the Mars 2020, presenting state estimation frameworks for both EDL and Ingenuity flights.

\vspace{1mm}
\subsubsection{The Lander Vision System (LVS)}

The framework presented in \cite{46186_2017} is a Map Relative Localization (MRL) algorithm deployed for Mars 2020 EDL that significantly reduces position errors that can be as large as 3200 m down to 40 m (from altitudes as high as 4200 m down to 2000 m) relative to a predefined landing site map, crucial for safer landings on Mars. However, LVS mission assumes a flat surface and map coverage of the landing site. Moreover, since MSH is equipped with on-board hazard avoidance \cite{9438289}, it does not strictly require safe landing sites to be identified on a map before like in the Mars 2020 mission. 

\vspace{1mm}
\subsubsection{MAVeN}

The navigation algorithm for NASA's Mars Helicopter~\cite{doi:10.2514/6.2019-1411} implements an Extended Kalman Filter offering several advantages over other vision-based velocimetry methods. It utilizes a relatively low-order 21-state filter for comprehensive 6-DOF pose estimation and can maintain a stable hover condition. It additionally integrates altimeter measurements for scale observation in the context of the Ingenuity Mars Helicopter navigation. However, MAVeN requires accurate horizontal position prior to determine feature depth from a 3D surface. 

\subsection{Range-Visual-Interial Odometry for Robotics}

The work done in~\cite{DBLP:journals/ral/DelauneBB21},~\cite{7989603} and~\cite{6696807}, shows that scale observability becomes a major concern under conditions of zero excitation of the IMU in monocular visual-inertial odometry frameworks, as scale error leads to position and velocity drift that can drastically affect navigation performance. This emphasizes the need for additional sensors like altimeters to address scale observability concerns under uniform motion. 

Modern 1D Laser Range Finder (LRF) devices come in compact, lightweight, and power-efficient packages, making them suitable in particular for robots with limited resources. The work of~\cite{DBLP:conf/icra/FuSM19} integrates a downward-facing 1D-LRF within an Extended Kalman Filter structure, expanding the state space with the distance from the ground state variable to enhance localization performance. Alternatively~\cite{f3858f712432459ba4f782b57629ef9c} implements a topographic SLAM algorithm relying on a single altimeter sensor and a rectangular panel map structure that is updated and refined at every iteration. Two other promising solutions,~\cite{wang2022revio} and ~\cite{DBLP:journals/corr/abs-2102-13406}, integrate event cameras and downward looking LRF sensors to improve the accuracy of position estimation by constructing additional range constraints and demonstrate higher accuracy, especially in fast-motion, low-light and highly dynamic scenes. While all the aforementioned approaches leverage 1D-LRF data to address scale observability challenges and improve the performance of the estimator, it is noteworthy that they share a common underlying assumption — they tend to assume a global planar structure for the terrain where visual features are detected. This assumption poses limitations in scenarios with diverse and non-planar terrains, as encountered in MAHD.

\vspace{1mm}

\begin{figure}[ht]
    \centering
    \includegraphics[width=0.35\textwidth]{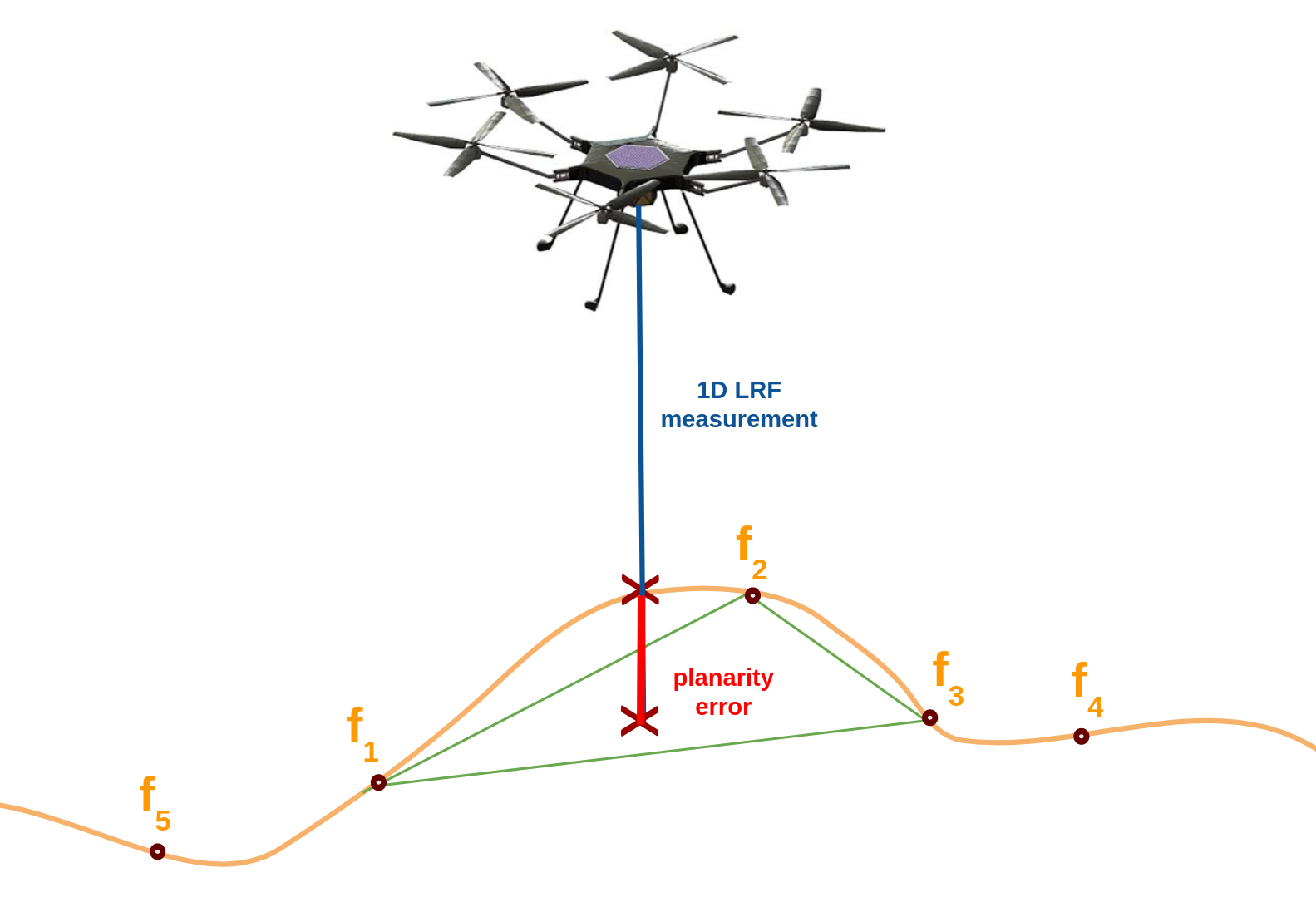}
    \caption{Schematic representation of the local planarity terrain assumption violation in the range-facet model in xVIO.}
    \label{img:planarity_error_scheme}
\end{figure}

The work done in~\cite{DBLP:journals/ral/DelauneBB21},~\cite{DBLP:journals/corr/abs-2010-06677} relaxes the aforementioned scene assumption from globally-flat to locally-flat by leveraging LRF measurements in a range-visual-inertial odometry implementation to make the scale observable. The method constrains the depth of the visual features estimated by VIO to the accurate LRF measurements. The method unfolds by partitioning the scene into triangular facets through a Delaunay triangulation of the features and assuming local flatness of each one of them. The range measurement is then expressed as a nonlinear function of the features corresponding to the 3 vertices of the triangle in which the
intersection of the LRF beam with the scene takes place, and it is used for the EKF update. The primary challenge associated with this approach in the context of MAHD arises from the relatively slow dynamics of the system during EDL compared to the high distance from the ground, while flying on rough terrain environments. In such conditions, the local planar assumption is highly violated due to the fact that the region encompassed by the facet extends over a terrain area that is far from being planar, as illustrated in Fig.~\ref{img:planarity_error_scheme}. Consequently, this leads to a higher degree of error being introduced into the navigation pipeline, ultimately resulting in a degraded performance of the state estimator.

In this paper, we delve into a critical problem: the observation of the scale in VIO under the constraints of constant-velocity motion and across any scene structure. To the best of our knowledge, no VIO method currently available offers a means to incorporate 1D-LRF measurements without necessitating assumptions about the terrain type.

\begin{figure}[ht]
     \vspace{1mm}
     \centering
     \includegraphics[width=0.3\textwidth]{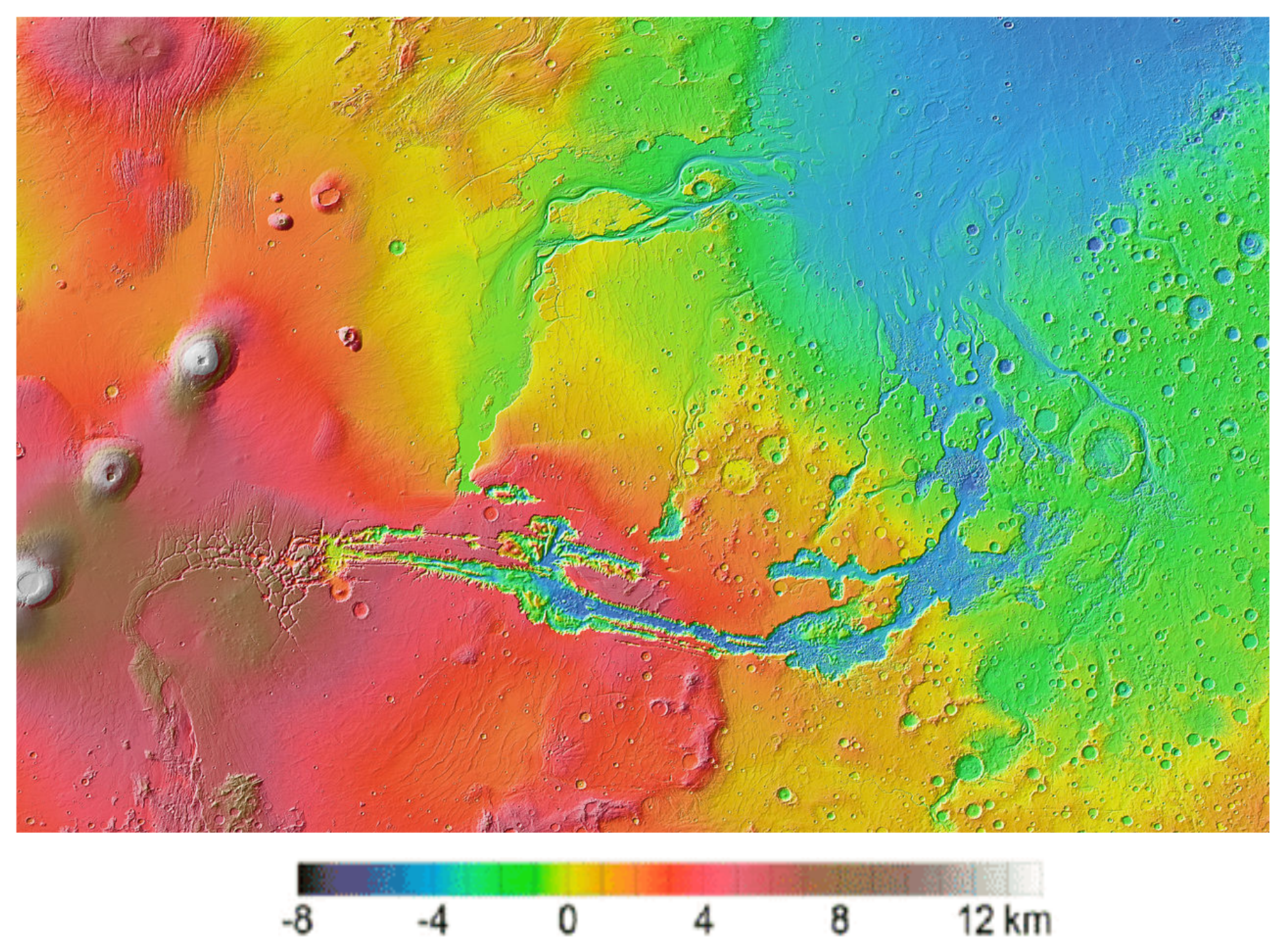}
     \caption{Elevation map of Valles Marineris from \cite{NASA-JPL:valles_marineris_elevation}. The elevation heat map shows that some areas in the site present abrupt changes in elevation, going from $-4000$ to $+4000$ m measured from the Mars Orbiter Laser Altimeter (MOLA).}
     \label{img:valles_marineris_elevation_map}
\end{figure}

\section{Terrain-Invariant Range-VIO}
\label{sec:method}

This paper's key contribution lies in integrating 1D-LRF measurements into the visual-odometry framework of the MSH, without assuming terrain planarity. In this section we first provide an overview of the xVIO architecture and then we introduce an innovative method addressing the aforementioned challenges, ensuring enhanced navigation performance for the MSH.

\subsection{IMU states and propagation}

The architecture of the xVIO framework~\cite{DBLP:journals/ral/DelauneBB21} is based on a tightly-coupled Extended Kalman Filter (EKF) formulation, that fuses together visual and range updates with inertial state propagation. The state vector $x$ of the EKF is divided between states related to the IMU $x_I$ and those related to vision $x_V$. The inertial states are the following
\begin{equation}\label{eq:state_vector}
    x_{I} = \left[ {p_{w}^{i}}^T \hspace{1.5mm} {v_{w}^{i}}^T \hspace{1.5mm} {q_{w}^{i}}^T \hspace{1.5mm} {b_{g}}^T \hspace{1.5mm} {b_{a}}^T \right]^T
\end{equation}
and they include position, orientation and velocity of the IMU frame with respect to the world frame, and the gyroscope and accelerometer biases. The state estimate is propagated over time through the IMU measurements, together with the corresponding subblocks of the error covariance matrix to first order. More details can be found in \cite{DBLP:journals/corr/abs-2010-06677}.

\subsection{Feature tracking and EKF-SLAM update}

Visual features are used in xVIO to construct a filter update following the hybrid SLAM-MSCKF paradigm~\cite{6577998}. Corners are detected in the images using the FAST algorithm and tracked using the pyramidal implementation of the Kanade-Lucas-Tomasi tracker. SLAM features represent the visual features defining the subset $x_F$ of the vision states $x_{V} = \left[ {x_S}^T \hspace{1.5mm} {x_F}^T \right]\in \mathbb{R}^{7M+3N}$ that are used in the EKF formulation. The position $\prescript{c_i}{}{p_j}$ of a SLAM feature $f_j$ is parameterized with respect to the camera frame $\{ c_i \}$ at time $i$ and then linearized to be used in the EKF visual update. Being $z_j$ the image measurement of feature $p_j$ in the normalized plane $\prescript{c_i}{}{z} = 1$ of camera $\{ c_i \}$, the measurement model for the SLAM visual update can be expressed as follows 
\begin{equation}
    \prescript{i}{}{z_j} = \frac{1}{\prescript{c_i}{}{z_j}} 
        \begin{bmatrix}
           \prescript{c_i}{}{x_j} \\
           \prescript{c_i}{}{y_j}
    \end{bmatrix} + \prescript{i}{}{n_j}  
\end{equation}
where $ \prescript{c_i}{}{p_{j}} = \begin{bmatrix} \prescript{c_i}{}{x_j} & \prescript{c_i}{}{y_j} & \prescript{c_i}{}{z_j}  \end{bmatrix}^T = C(q_{w}^{c_i}) (p_{w}^{j} - p_{w}^{c_i}) $ and $\prescript{i}{}{n_j}$ is a zero-mean Gaussian measurement noise with covariance $\prescript{i}{}{R_j} = \sigma_V^2 I_2$. Expressing the measurement model as a function of the state defines a nonlinear visual measurement function that can be represented as follows
\begin{equation}\label{eq:EKF_measurement_function}
    \prescript{i}{}{z_j} = h(p_{w}^{{c_{i}}_{j}}, q_{w}^{{c_{i}}_{j}}, \prescript{w}{}{p_j}) + \prescript{i}{}{n_j}  
\end{equation}
The direct measurement of $\prescript{i}{}{z_j}$ comes from any image point tracked over time by the visual front-end. The Cartesian coordinates of $p_{j}$ in world frame $\{ w \}$ can be expressed as
\begin{equation}\label{eq:EKF_SLAMupdate1}
    \prescript{w}{}{p_j} =     
    p_{w}^{{{c_{i}}_j}} + \frac{1}{\rho_{j}} C(q_{w}^{{{c_{i}}_j}})^{T}
    \begin{bmatrix}
           \alpha_{j} \\
           \beta_{j} \\
           1
    \end{bmatrix} - q_{w}^{c_{i}}
\end{equation}    
where $\{ {{c_{i}}_j} \}$ represents the anchor camera frame of $p_j$. In this work, SLAM features are initialized with a planar feature depth initialization module and semi-infinite depth uncertainty \cite{4637878}. The inverse feature depth is computed from the distance between the current estimate of the drone and a plane defined at runtime from the pose prior. 

On the other hand, the MSCKF features are associated with the sliding window states $x_S = \left[ {p_{w}^{c_{1}}} \hspace{0.3mm} ... \hspace{1.2mm}  {p_{w}^{c_{M}}} \hspace{1mm}  {q_{w}^{c_{1}}} \hspace{0.3mm} ... \hspace{1.2mm}{q_{w}^{c_{M}}} \right] \in \mathbb{R}^{7M}$ that include the positions $\{ p_{w}^{c_{i}} \}$ and the orientations $\{ q_{w}^{c_{i}} \}$ of the camera frame in the last $M$ image time instances. The image coordinates of the MSCKF features that are tracked over time are triangulated to determine the 3D camera pose at time $i$, which is then linearized for the update. 
Given the focus on EDL trajectories starting at 12000 m above the ground, the MSCKF module is not used in this work. At elevations as high as those encountered in MAHD and velocities similar to those prescribed for the maneuver, the triangulation of features over time pose challenges due to small baselines. Consequently, for our examination, we depend solely on the xVIO SLAM module, limiting the SLAM features to $N = 15$ in order to align with the computational constraints that apply to the MSH.

\subsection{Online range-features initialization with 1D-LRF measurements}

This section introduces a novel method to integrate altimeter readings into xVIO without making any type of assumption about the terrain. Without generalization loss, the approach assumes the alignment of the 1D-LRF frame with the camera's frame. The method consists in tracking the corner score of the central pixel over time, and whenever its value exceeds a predefined threshold, a new visual feature that we refer to as \textit{ranged-feature} is defined and incorporated into the other SLAM features used in the EKF visual update. Unlike other features, the depth of the range-feature is precisely known from the 1D-LRF, allowing for initialization with an exact measurement and low variance. The method leverages the accurate depth information to improve the state estimation through covariance propagation in the EKF.

The corner score that we consider to trigger new range-features is the minimum eigenvalue metrics defined in \cite{Bouguet2000}. This metrics is the same used on a lower level by the Lucas-Kanade optical flow tracker in the xVIO framework. For this reason why we do not rely on the FAST detection score, as we want to include those features that are more likely to be tracked over longer period. Given an image $I(x, y)$ and its image derivatives $I_x(x, y)$ and $I_y(x, y)$, the spatial gradient matrix defined in a windows of size $[-w_x, w_x] \times [-w_y, w_y]$ around an image point of coordinates $(x_{LRF}, y_{LRF})$ is defined as follows
\begin{equation} \label{eq:spatial_gradient_matrix}
    G = 
    \sum_{x={x_{LRF}-w_x}}^{x_{LRF}+w_x} \hspace{1mm} \sum_{y={{y_{LRF}}-{w_y}}}^{{y_{LRF}}+{w_y}} 
    \begin{bmatrix}
        I_x^2 & I_x I_y  \\
        I_x I_y & I_y^2 \\
    \end{bmatrix}
\end{equation}
The metrics is defined as the minimum eigenvalue of the spatial gradient matrix $G$. A new range-feature is determined when both of the following conditions are satisfied:

\begin{enumerate}
    \item The minimum eigenvalue score is above a predefined threshold;
    \item The score value at time $t$ is greater than the score computed in the subsequent N iterations, with N being a parameter specified at run-time.
\end{enumerate}

\begin{figure}[ht]
   \centering
   \includegraphics[width=0.36\textwidth]{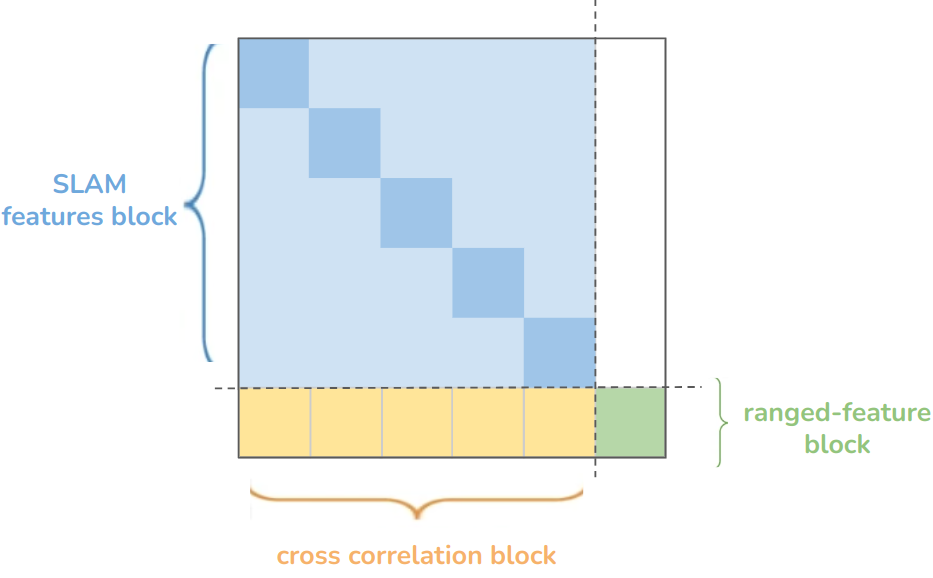}
   \caption{Block representation of the feature state covariance matrix. Through their cross-correlation block with the single ranged-feature, the depth estimates of the unranged feature converges to the correct metric depth under a large-enough translation motion. This is equivalent to a triangulation.}
   \label{img:covariance_block_matrix}
\end{figure}

\begin{figure}[ht]
     \centering
     \includegraphics[width=0.38\textwidth]{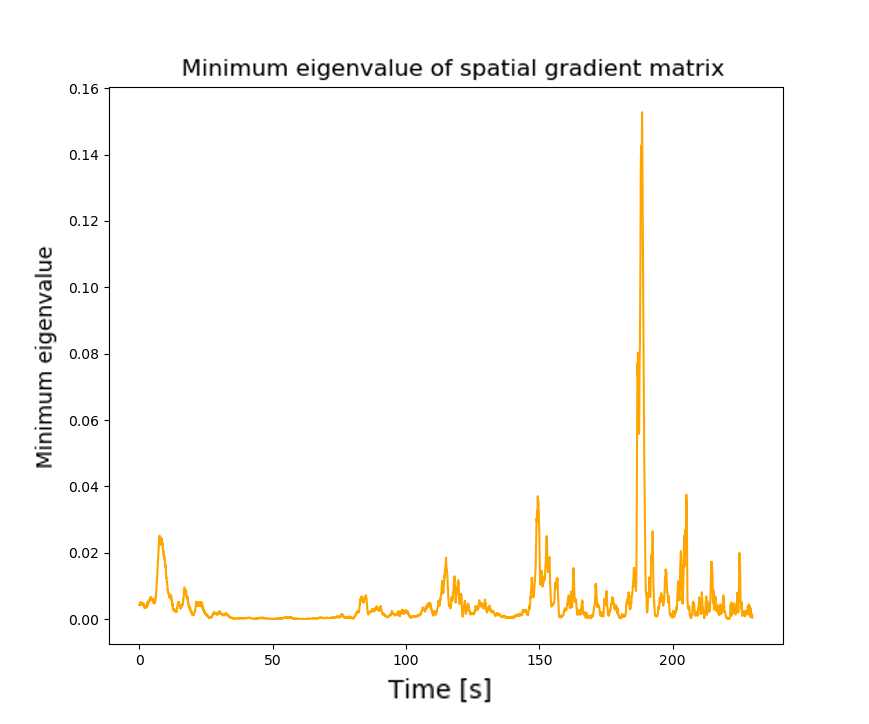}
     \caption{ Minimum eigenvalue of the spatial gradient matrix computed in a $9\times9$ window around the central pixel.}
     \label{img:minimum_eigenvalues_range-feat}
\end{figure}

Condition (1) serves the purpose of selecting good features to track, aiming to focus the attention on image points exhibiting substantial variations in intensity or texture, such as corners. By filtering out points with low eigenvalues, we effectively prioritize features that are more likely to yield meaningful information and can be tracked over time. Condition (2) defines an interval of size N in which we want the score to be monotonically decreasing. Ideally, the peak represents a local maximum of the curve after which the values decrease. This condition is important to avoid choosing sub-optimal pixels for a representative visual feature. The plot in Fig.~\ref{img:minimum_eigenvalues_range-feat}  shows the corner score over time in a sample MAHD-like trajectory. Distinct peaks are discernible in the plot, each corresponding to favorable features suitable for tracking. Once a pixel satisfies the aforementioned conditions, it becomes a candidate for inclusion in the feature matches and subsequently undergoes tracking, following a protocol similar to that of any other SLAM feature. A visual representation of the method is shown in Fig.~\ref{img:representation_range_feature_initialization}. As it persists in being tracked for a duration exceeding the minimum track length parameter specified at run-time, it attains eligibility for inclusion in the set of SLAM tracks used for updates. The feature state associated with the new range-feature $ f_{{LRF}_j} = [ \alpha_{{LRF}_j}, \beta_{{LRF}_j}, \rho_{{LRF}_j} ] ^ T $ represents the \textit{inverse-depth parameterization}\footnote{The inverse-depth parameterization has been used to represent feature coordinates in SLAM due to its improved depth convergence properties \cite{4637878}.} of the measurement $\prescript{c_i}{}{p_{{LRF}_j}}$ expressed with respect to camera frame {$c_i$} at time $i$, with $\alpha_{{LRF}_j}$ and $\beta{{LRF}_j}$ being the normalized image coordinates of the feature over time. Being $\prescript{i}{}{\widetilde{z}_r}$ the value the altimeter measures at time $i$, the initial inverse-depth estimate of the new feature is set to $\hat{\rho}_{0_{{LRF}_j}} = \frac{1}{\prescript{i}{}{\widetilde{z}_r}}$, with the corresponding standard deviation $\sigma_{0_{{LRF}_j}}$ being initialized with the standard deviation of the 1D-LRF model used in practice. Whenever a new range-feature is being tracked, the state space estimate is augmented in the following way
\begin{equation}\label{eq:range-feat-initialization}
    \hat{x}_{aug_{k|k}} \leftarrow 
    \begin{bmatrix}
      \hat{x}_{k|k} \\
      \hat{\alpha}_{0_{{LRF}_j}} \\
      \hat{\beta}_{0_{{LRF}_j}} \\
      \hat{\rho}_{0_{{LRF}_j}} \\
    \end{bmatrix}
\end{equation}
Once the range-feature has been added to the SLAM tracks, the update rule for that feature follows the same pattern of any other SLAM feature. The fundamental principle underlying this implementation is the expectation that the accurate depth information of a particular range-feature should propagate through the covariance in the EKF update, following the scheme presented in Fig.~\ref{img:covariance_block_matrix}. By doing so, it not only refines the estimation of that feature but also contributes to the overall improvement of other SLAM features.

\section{Experimental setup}
\label{sec:results}

Synthetic datasets were generated to assess the efficacy of the implemented method on the xVIO framework. These datasets comprise sequences of in-flight camera images, IMU and altimeter readings with representative sensor noise, accompanied by full-state ground truth information including pose. These datasets were used to compare our approach to the state-of-the-art in an extensive Monte Carlo analysis.

\subsection{Representative Mars Simulation Environment}

To ensure a robust evaluation of the proposed algorithm, key requirements include a rendering engine able to generate high-fidelity images of 3D terrain models captured at distances as high as 12000 m and a comprehensive sensor suite capable of logging measurements during simulation.

\subsubsection{Addressing limitations in existing simulators}

Existing drone simulators, such as Airsim~\cite{10.1007/978-3-319-67361-5_40}, face limitations when it comes to precisely simulating the specific MAHD requirements. 
Specifically, initiating a drone on a constant-velocity trajectory from a stationary state poses challenges. This is because the drone's dynamics necessitate a certain level of acceleration to begin movement, inevitably causing some degree of IMU excitation. Consequently, achieving entirely steady constant-velocity trajectories without any form of IMU excitation is unattainable within these constraints. This requirement stems from the need of replicating representative and critical phases of MAHD like the parachute phase\footnote{In the parachute phase the primary goal is to gradually decelerate the MSH and the jetpack system until they reach a state of zero acceleration.} and the jetpack-hovering phase, where near-zero acceleration is attained.

\begin{figure}[ht]
   \vspace{1mm}
  \centering
  \begin{subfigure}{0.48\textwidth}
    \includegraphics[width=\linewidth]{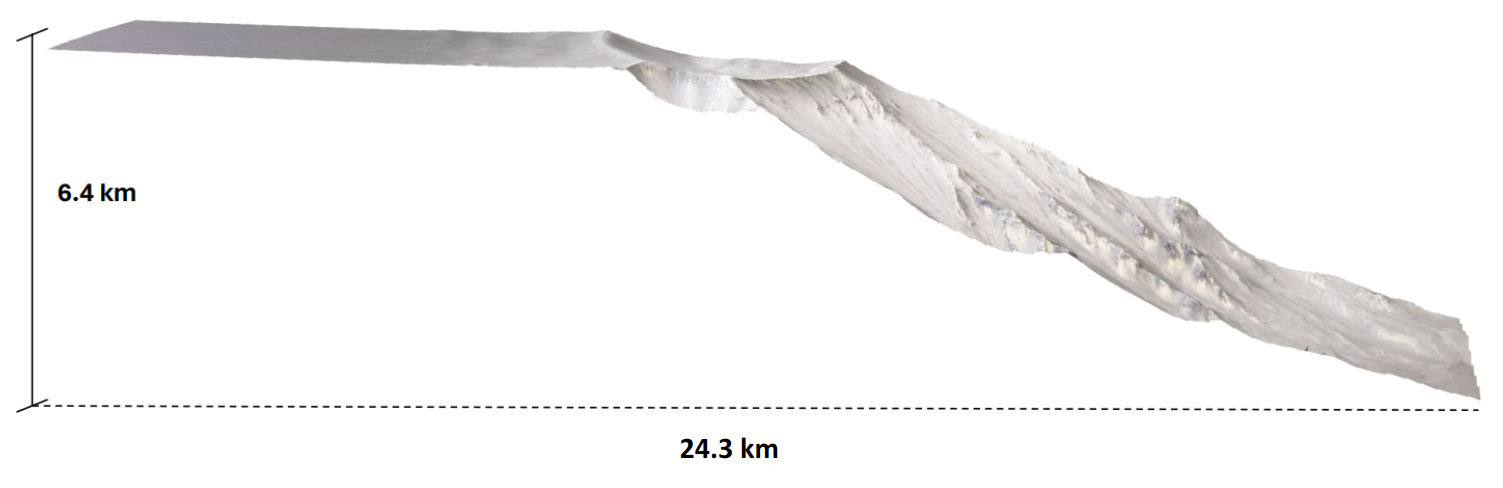}
    \label{img:HiRISE_3D-model_sim_2ß}
  \end{subfigure}

  \centering
  \begin{subfigure}{0.25\textwidth}
    \includegraphics[width=\linewidth]{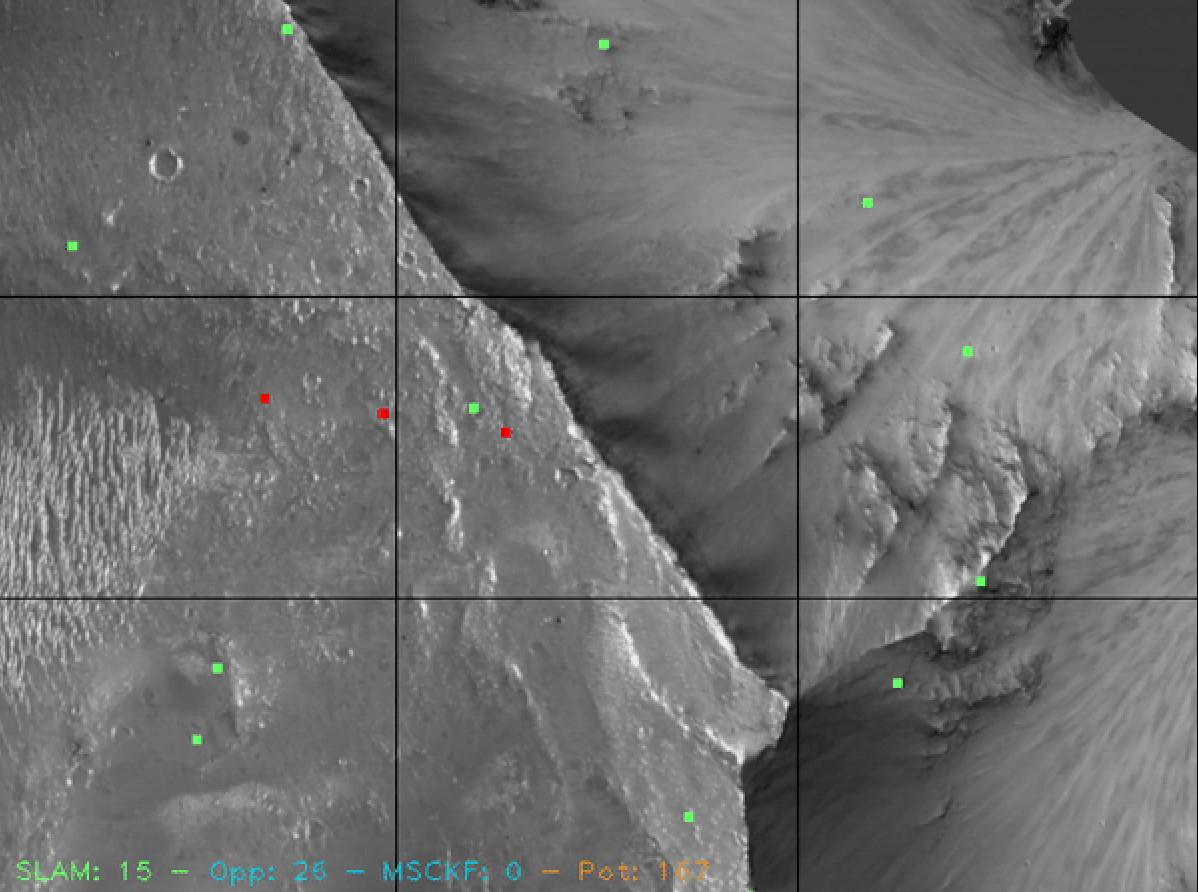}
    \label{img:HiRISE_3D-model_sim_1}
  \end{subfigure}
  
  \caption{The upper image shows a lateral view of a site in Valles Marineris, generated from a HiRISE DTM rendered in the representative simulation environment. The image below shows a camera image view on the model while running xVIO. Green dots correspond to SLAM features, while red dots to ranged-features. }
  \label{img:HiRISE_3D-model_sim}
\end{figure}

A custom simulation environment was designed by seamlessly integrating external packages for trajectory generation \cite{rtb} into the Airsim drone simulator \cite{10.1007/978-3-319-67361-5_40} on Unity. This integration facilitates the generation of straight constant velocity trajectories and enables the decoupling from the drone dynamics during the data generation process. 

\subsubsection{Mars environment model generation}

In order to achieve the highest level of precision in the custom simulation environment, we use Digital Terrain Models (DTMs) of various Martian sites sourced from the High-Resolution Imaging Science Experiment (HiRISE) website. HiRISE data are derived from measurements taken by an orbiting spacecraft equipped with a powerful camera capable of capturing images covering extensive Martian terrain while discerning features with a resolution of about 1 meter. The process takes place with the integration of DTMs into Blender, facilitated by a specialized add-on tailored for HiRISE DTM integration. An example of a 3D model rendered in the simulation environment can be observed in Fig.~\ref{img:HiRISE_3D-model_sim}.

\subsection{Sensor noise specifications}

To ensure uniformity and reproducibility in all experiments, a consistent sensor setup was employed. The simulated IMU, modeled after the \textit{MPU-9250} model, features a gyro noise spectral density of $0.0013 \hspace{1mm} rad/s/\sqrt{Hz}$, a gyro bias random walk of $0.00013 \hspace{1mm} rad/s^2/\sqrt(Hz)$, an accelerometer noise density of $0.0083 \hspace{1mm} m/s^2/\sqrt{Hz}$ and an accelerometer bias random walk of $0.00083 \hspace{1mm} m/s^3/\sqrt{Hz}$. The deliberate use of the lower-quality \textit{MPU-9250} IMU model, instead of the planned \textit{STIM-300} for the MSH, aimed to test the navigation system under more challenging conditions. This approach simulated potential vibrations and disturbances in a degraded scenario, creating a safety margin for real-world challenges. The altimeter model employed for experiments was the \textit{DLEM-30} Laser Range Finder, incorporating a 1m standard deviation noise component on top of true altitude values. The \textit{DLEM-30} is a high-precision LRF with a measurement range from 10m to 14000m and sub-meter accuracy. The camera model, integrated into the Unity simulation environment, featured a horizontal field of view of 90 deg and generated images at a resolution of 640 × 480 pixels.

\begin{figure}[ht]
   \centering
   \includegraphics[width=0.28\textwidth]{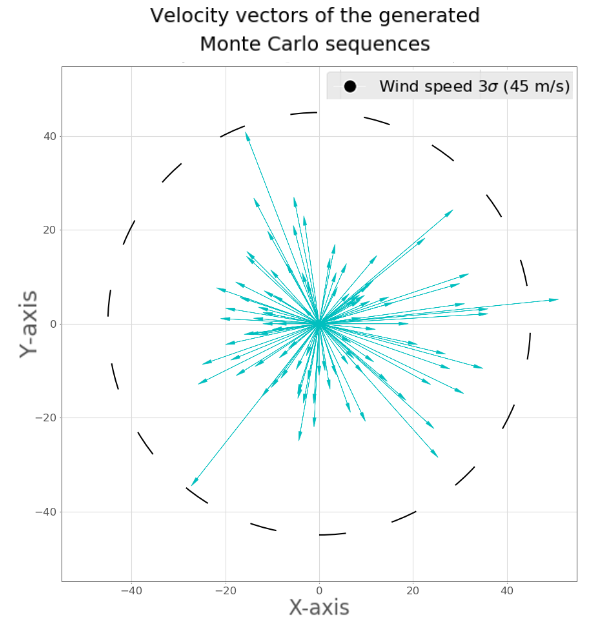}
   \caption{Velocity vectors of the generated sequences for the Monte Carlo analysis. The vectors represent the errors on the x and y components of the velocity due to lateral wind.}
   \label{img:velocity-vectors_MonteCarlo}
\end{figure}

\subsection{Monte Carlo analysis setup}

A Monte Carlo analysis is conducted to provide a more rigorous evaluation of the algorithm's performance within the context of the simulation environment by systematically taking into account possible uncertainties representative of real-life scenarios encountered during the deployment of the MSH on Mars. 
Firstly, trajectories with different initial and final positions are sampled within the environment, while ensuring both a consistent landing altitude of 200 meters above the ground and visibility of the 3D environment in the camera FOV at every step. Initial velocity parameters are perturbed, introducing a 45 m/s $3\sigma$ error exclusively to the x and y components coming from the horizontal winds distribution models available for Mars \cite{9843825}, since it is the maximum velocity at which the parachute can be dragged sideways with respect to the terrain. The z-component of the velocity is instead firmly set at the terminal velocity, directed downwards at approximately -56 m/s, because the vertical wind distribution is close to zero and we do not need to add any variability there. Additionally, image noise, in the form of shot noise with a 1-pixel $1\sigma$ standard deviation, is injected into the simulated image data. The IMU and LRF models adhere to previously outlined specifications, namely those of the \textit{MPU-9250} IMU and \textit{DLEM-30} altimeter, respectively.

\vspace{1mm}

On the evaluation front, further sources of uncertainty are introduced. The initial velocity provided during state estimator initialization remains fixed at [0, 0, -56] $m/s$, despite the real value of the x and y velocities is different at each trajectory. An additional layer of complexity is introduced through the imposition of an initial attitude error, perturbing the true configuration by a 1 deg $3\sigma$ error across the x, y, and z axes. In addition to that, an error of the $20\%$ on the true initial position of the drone is set at run-time. This Monte Carlo analysis seeks to analyze the algorithm's performance across a spectrum of real-world scenarios, providing invaluable insights into its reliability.

\section{Results}
In this section, we first analyze the limitations of the range-facet model in the scope of MAHD and then we evaluate the effectiveness of the novel range-VIO framework on representative trajectories presented earlier. The plots we are presenting focus solely on velocity, as it is the main objective during the descent phase, crucial for jetpack control.

\begin{figure}[ht]
    \begin{subfigure}[b]{0.24\textwidth}
        \centering
        \includegraphics[width=1.0\textwidth]{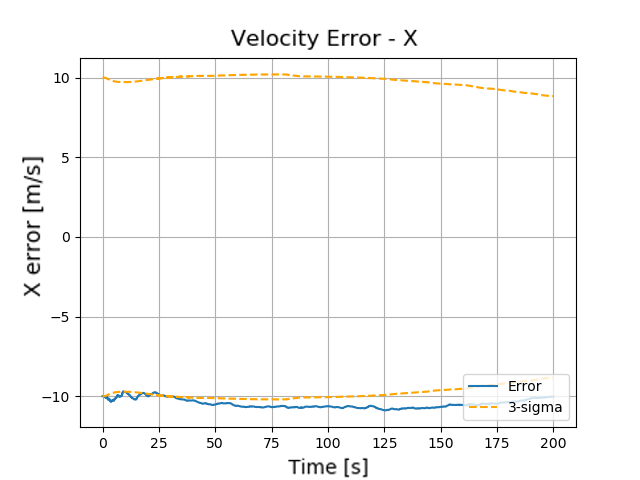}
        \caption{}
    \end{subfigure}
    \hfill
    \begin{subfigure}[b]{0.24\textwidth}
        \centering
        \includegraphics[width=1.0\textwidth]{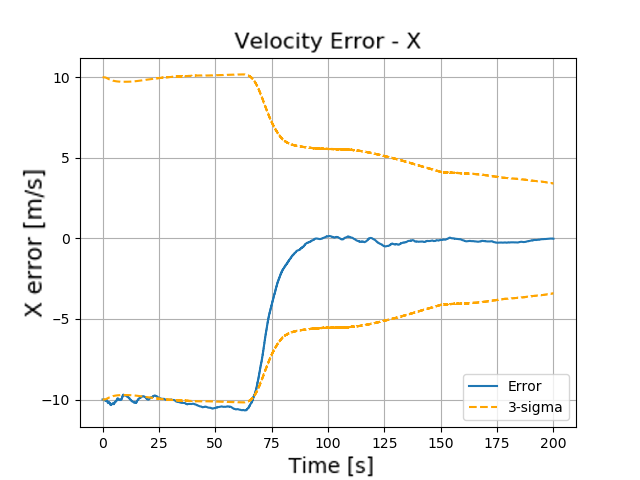}
        \caption{}
    \end{subfigure}
    \caption{Comparison between velocity error from the estimated state on the x axis (direction of motion) (a) without and (b) with range-feature initialization method.}
    \label{img:results_scale-observ}
\end{figure}

\subsection{Failure mode in xVIO range-facet model}

\begin{figure}[ht]
  \centering
  \begin{subfigure}{0.485\textwidth}
    \includegraphics[width=1.0\linewidth]{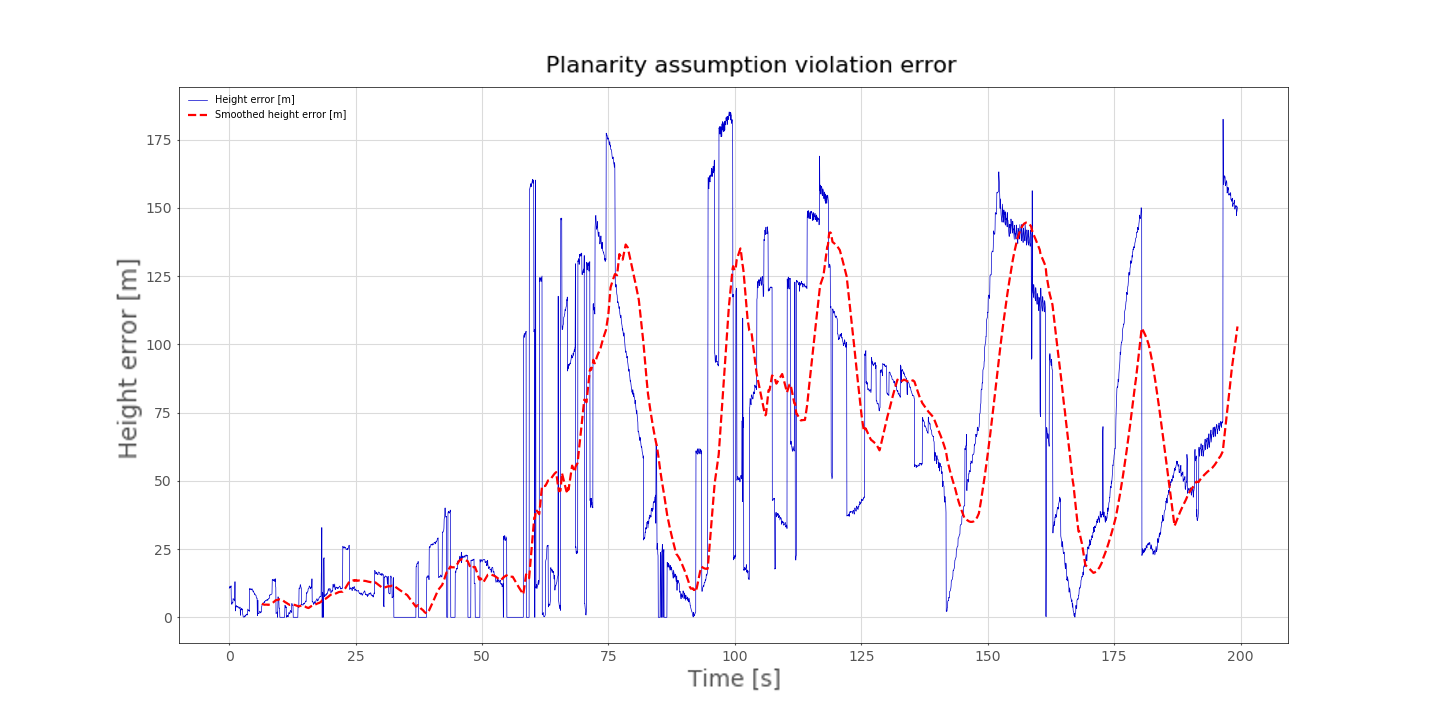}
    \caption{}
    \label{img:planarity_error_plot}
  \end{subfigure}

  \centering
  \begin{subfigure}{0.485\textwidth}
    \includegraphics[width=1.0\linewidth]{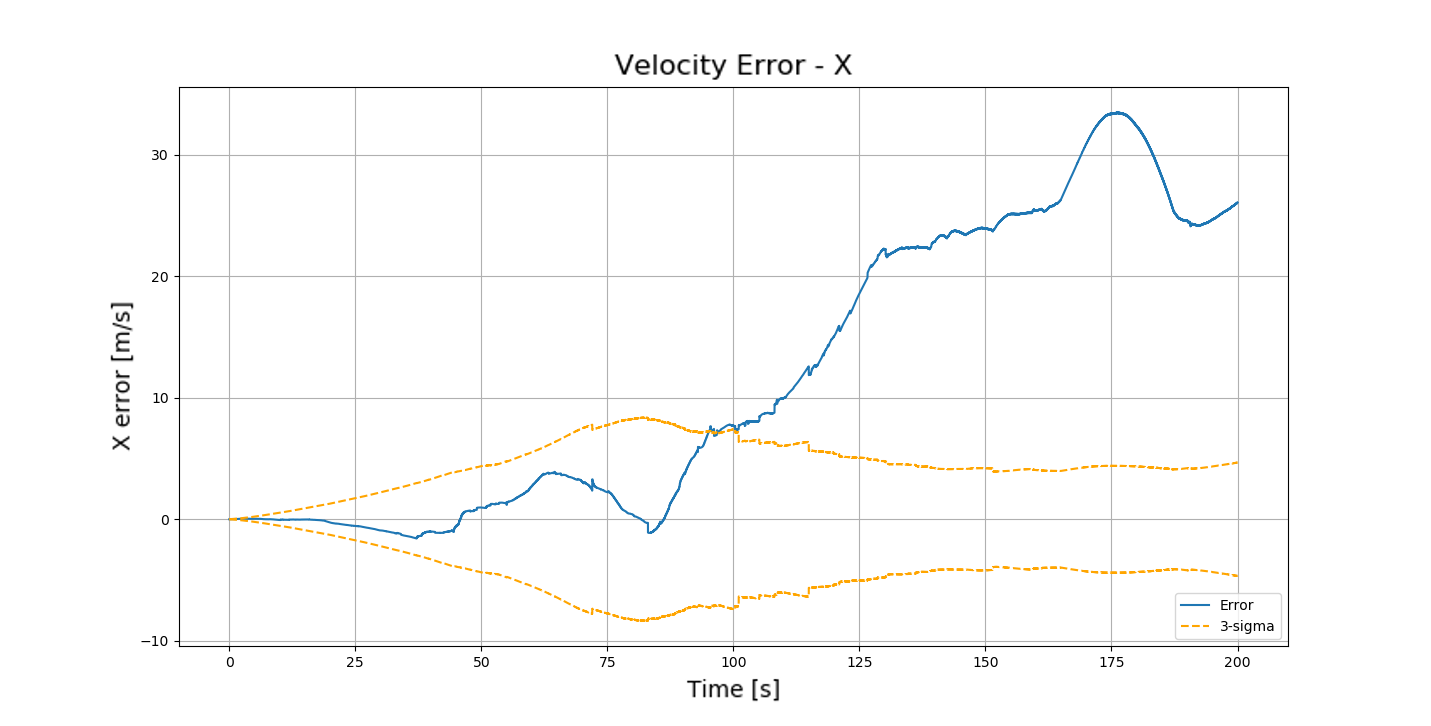}
    \caption{}
    \label{img:planarity_error_RANGE}
  \end{subfigure}
  
  \caption{Comparison between (a) error between true altimeter value and projected height of the facet plane (b) velocity error on X-axis of xVIO with range-facet update evaluated on a horizontal trajectory. These plots show that the state-of-the-art method fails as the facet model assumption is violated.}
  \label{img:planarity_error_violation}
\end{figure}

The primary challenge associated with the range-facet model approach in the context of MAHD arises when operating at high altitudes and in rough terrain environments. In such conditions, the local planar assumption is highly violated due to the fact that the region encompassed by the facet extends over a terrain area that is far from being planar. Consequently, this leads to a higher degree of error being introduced into the navigation pipeline, ultimately resulting in a degraded performance of the state estimator. To validate the previous hypothesis, xVIO with range-facet update has been evaluated on a MAHD sequence generated within an image-based simulation environment on a terrain presenting significant height variations. The evaluation results shown in Fig.~\ref{img:planarity_error_violation} reveal that the performance of the estimator directly correlates to the violation of the local planarity of the central facet when compared to the real 3D structure of the terrain.

\subsection{Scale observability with range-feature initialization}

One of the primary objectives of including range-features in the pipeline is to assess the algorithm's effectiveness in addressing scale observability. The evaluation involves a simplified horizontal trajectory with a representative maximum velocity vector of [50, 0, 0] m/s at a constant elevation of 12000 m above a uniformly flat terrain, isolating effects from the 3D complexity of the environment. Initial condition errors, including a 20\% error in position and velocity vector norm, introduce a 20\% overall scale error. Results shown in Fig.~\ref{img:results_scale-observ} demonstrate that without range-feature initialization, the scale remains unobserved. However, after triggering and incorporating the first range-feature into the SLAM update process, a notable transformation occurs and the velocity error on the x axis goes to a value close to zero. While observing the scale in terrain-relative navigation does not entirely nullify position errors, it effectively brings the velocity error close to zero.

\begin{figure*}
    \begin{subfigure}[b]{0.325\textwidth}
        \centering
        \includegraphics[width=\textwidth]{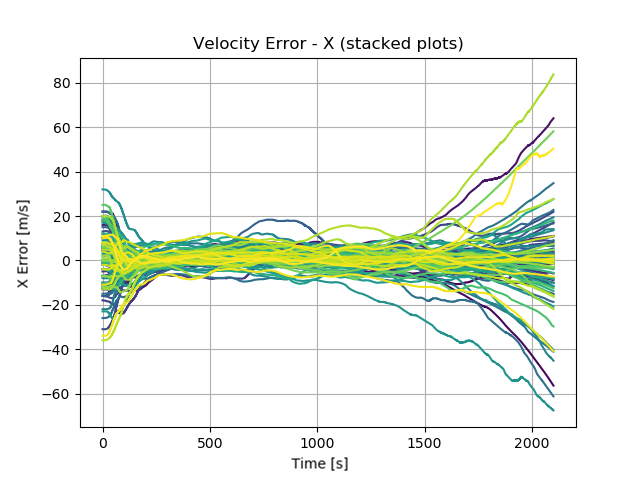}
        \caption{}
        \label{img:montecarlo_vision_velocities}
    \end{subfigure}
    \hfill
    \begin{subfigure}[b]{0.325\textwidth}
        \centering
        \includegraphics[width=\textwidth]{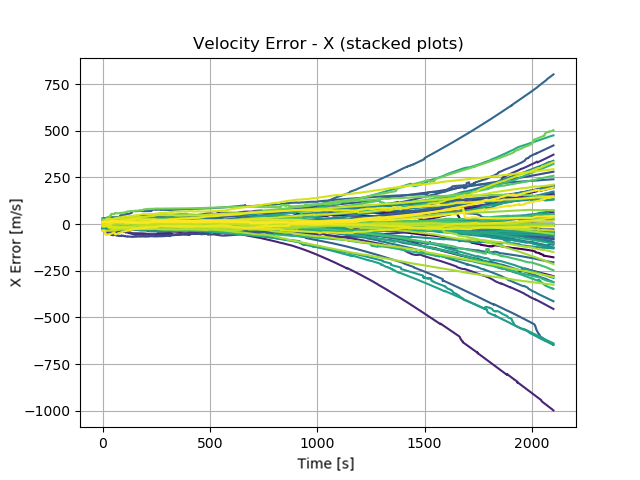}
        \caption{}
        \label{img:montecarlo_facet_velocities}
    \end{subfigure}
    \hfill
    \begin{subfigure}[b]{0.325\textwidth}
        \centering
        \includegraphics[width=\textwidth]{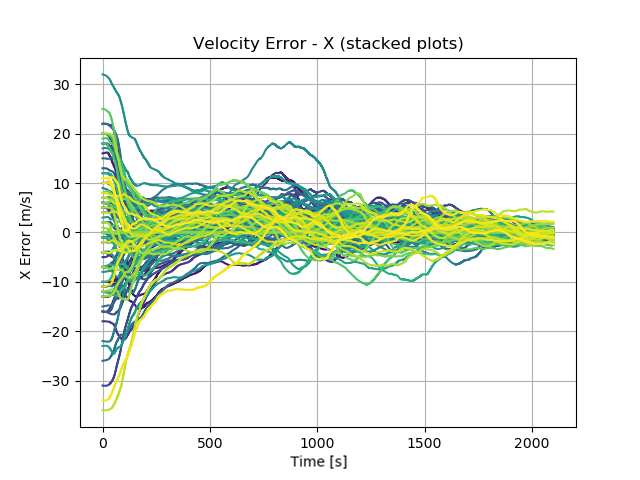}
        \caption{}    
        \label{img:montecarlo_range-init_velocities}
    \end{subfigure}
    \caption{Comparison of the velocity error plots of 100 generated MAHD-like sequences with the Monte Carlo simulation setup (stacked visualization) being evaluated in (a) xVIO with visual EKF update (b) xVIO with range-facet update (c) xVIO with range-features initialization. Velocity is the main metrics we monitor for stable jetpack control during MAHD.}
    \label{img:montecarlo_velocities_error}
\end{figure*}

\subsection{Monte Carlo analysis on MAHD trajectories}

A Monte Carlo evaluation has first been conducted employing the xVIO framework with exclusively SLAM-based visual updates. Subsequently, the analysis encompasses the incorporation of the range-facet update model, followed by the activation of range-feature initialization. The results of the evaluation are shown in Fig.~\ref{img:montecarlo_velocities_error}, focusing solely on velocity error plots along the x-axis. The behavior on the y and z axes remains consistently the same.

The results of the Monte Carlo analysis on xVIO with only visual-inertial updates, as shown in Fig.~\ref{img:montecarlo_vision_velocities}, unveil a divergence issue in certain trajectories. This divergence appears to stem from the erroneous initial position of the estimator, initialized with a 20\% error, which directly affects the SLAM features depth initialization, being it sensitive to the initial pose prior and the current state estimate. The erroneous pose prior leads to a cascading effect on the initialization of the inverse-depths of each new feature during the SLAM features initialization process, ultimately leading to the divergence observed in some trajectories. 
Moving to xVIO with the incorporation of the range-facet EKF update, the Monte Carlo analysis displayed in Fig.~\ref{img:montecarlo_facet_velocities} reveals divergence in nearly all individual runs. The primary culprit behind this inefficiency lies in the significant error associated with the local planarity assumption embedded in the method, rendering it ineffective within the context of MAHD. 
In contrast, the Monte Carlo analysis on xVIO using the range-feature initialization method showcases superior performance among the considered approaches. Deployed alongside a planar depth initialization of other SLAM features based on LRF measurements, this method proves effective in the context of MAHD. Unlike xVIO with visual updates only, the state estimator does not exhibit divergence issues with range-feature initialization.

\section{CONCLUSIONS}
\label{sec:conclusions}

This research extends the NASA JPL xVIO library with a range-features initialization module and performs an extensive evaluation of the framework under the challenging conditions of the Mid-Air Helicopter Delivery. To the best of our knowledge, the work represents the first attempt to integrate 1D-LRF readings into a visual-odometry framework without the need for ground planarity assumptions and makes it adaptable to any type of 3D terrain structure. This novel approach effectively maintains the capability to estimate scale and mitigate error drift during constant-velocity motion, all without relying on pre-existing maps. 

\vspace{1mm}

For future work, the research will extend into several critical areas of investigation. One significant avenue we intend to explore involves assessing the performance of a delayed feature initialization block, which aims to enhance the initialization of the SLAM features' depth. The goal is to enhance the accuracy of the MAHD trajectory by minimizing error during the initialization of other SLAM features depths. Furthermore, we plan to conduct additional evaluations on the implemented method within a scenario more closely resembling the complexities of the EDL process. This will include evaluating the proposed framework on more intricate trajectories to gain a deeper understanding of its robustness and adaptability in diverse scenarios. In addition to simulated scenarios, we intend to validate the algorithm's performance in real-world operational conditions using collected data. These extensions will contribute significantly to expanding our knowledge of the algorithm's capabilities and identifying potential areas for optimization and refinement.

\section*{ACKNOWLEDGMENT}

This research was carried out at the Jet Propulsion Laboratory, California Institute of Technology, and was sponsored by JVSRP and the National Aeronautics and Space Administration (80NM0018D0004). ©2024. All rights reserved.

{\small
\bibliographystyle{IEEEtran}
\bibliography{all}
}

\end{document}